\documentclass[dvipsnames]{Interspeech}
\usepackage[utf8]{inputenc}
\usepackage[T1]{fontenc}

\interspeechcameraready
\title{Factors affecting the in-context learning abilities of LLMs \\for dialogue state tracking}

\author[affiliation={1,2}]{Pradyoth}{Hegde}
\author[affiliation={1}]{Santosh}{Kesiraju}
\author[affiliation={1}]{J\'{a}n}{\v{S}vec}
\author[affiliation={1}]{\v{S}imon}{Sedl\'{a}\v{c}ek}
\author[affiliation={1}]{Bolaji}{Yusuf}
\author[affiliation={1}]{Old\v{r}ich}{Plchot}
\author[affiliation={2}]{Deepak}{K T}
\author[affiliation={1}]{Jan}{\v{C}ernock\'{y}}

\affiliation{Speech@FIT}{Brno University of Technology}{Czechia}
\affiliation{}{Indian Institute of Information Technology Dharwad}{India}
\email{pradyothhegde@gmail.com, kesiraju@fit.vut.cz}

\usepackage{comment}
\usepackage{amsmath} 
\usepackage{graphicx}
\usepackage{tcolorbox}
\usepackage{multirow}
\usepackage{caption} 
\usepackage{tikz}
\usetikzlibrary{shapes.geometric, arrows}
\usepackage{xcolor}

\begin{document}

\maketitle

\begin{abstract}

    This study explores the application of in-context learning (ICL) to the dialogue state tracking (DST) problem and investigates the factors that influence its effectiveness. We use a sentence embedding based k-nearest neighbour method to retrieve the suitable demonstrations for ICL. The selected demonstrations, along with the test samples, are structured within a template as input to the LLM. We then conduct a systematic study to analyse the impact of factors related to demonstration selection and prompt context on DST performance. This work is conducted using the MultiWoZ2.4 dataset and focuses primarily on the OLMo-7B-instruct, Mistral-7B-Instruct-v0.3, and Llama3.2-3B-Instruct models. Our findings provide several useful insights on in-context learning abilities of LLMs for dialogue state tracking\footnote{Accepted to Interspeech'2025, ISCA}.
\end{abstract}

\section{Introduction}
Instruction-tuned large language models (LLMs) have demonstrated enhanced capabilities compared to traditional language models, enabling them to perform a broader range of tasks based on instructions \cite{ouyang2022:IT,chung2022:flanT5,wei2022finetuned}. A key capability of LLMs is in-context learning (ICL), which allows them to generalise to new tasks without requiring explicit fine-tuning, thus reducing the need for extensive task-specific training data and computational resources \cite{brown2020language,dong-etal-2024-survey,wu2023openicl}. This adaptability makes them particularly attractive for complex applications, such as task-oriented dialogue systems, where the ability to quickly adapt to new domains and user intents is highly desirable \cite{hu2022context}.

Task-oriented dialogue (TOD) systems are conversational systems designed to achieve specific, predefined goals, such as booking a flight, ordering food, or scheduling an appointment. To achieve these goals, the dialogue manager within the system maintains a representation of the current dialogue state. The dialogue state comprises the domain of the conversation (e.g. restaurant, taxi) and the relevant information or values associated with that domain, referred to as slots (e.g. restaurant-name, taxi-destination). In multi-turn dialogues, accurately tracking the domain and the corresponding slot values is known as dialogue state tracking (DST). This is crucial for the overall performance of TOD systems, as it directly impacts the system's ability to understand user requests, provide relevant information, and ultimately achieve the desired task completion \cite{balaraman2021recent}.

The effectiveness of ICL is heavily influenced by the selection of appropriate demonstrations \cite{liu2022makes,li-qiu-2023-finding,zhang-etal-2022-active,chang-jia-2023-data,peng-etal-2024-revisiting}. Several recent works have focused on improving demonstration selection strategies for LLMs \cite{chen-etal-2023-stabilized,li2023unified,shu2024comparative}. Rubin et al. \cite{rubin2022learning} explored a retrieval-based approach, training a model to score demonstrations based on LLM performance. Venkateswaran et al. \cite{venkateswaran2023district} fine-tuned a language model using ICL and observed that while fine-tuning can improve performance, the quality and diversity of the demonstrations remain critical factors, noting that low diversity and semantic similarity in the selected demonstrations can negatively impact performance. Interestingly, Min et al. \cite{Min2022Rethinking} found that the ground truth slot values in the demonstrations may not be as important as previously thought, as replacing them with random words did not significantly degrade performance, suggesting that the structure and context provided by the demonstrations may be more influential than the specific slot values themselves.

In this work, we study a nearest-neighbour-based retrieval to select relevant demonstrations (dialogue turns) for ICL-based DST.
These demonstrations are then formatted using a carefully designed prompt
template to effectively guide the LLM towards accurate state tracking. Specifically, we propose a modular prompt structure for DST, consisting of distinct blocks for the conversation history, the domain(s), and the corresponding slot values. This modularity allows for easy adaptation and experimentation with different prompt configurations.
This study focuses exclusively on in-domain examples. Due to the poor performance of zero-shot prompting in our format, we limit our investigation to demonstration-based DST.
We conduct a systematic investigation and identify the factors influencing both the demonstration retriever and the prompt structure in the final DST performance. Our investigation reveals that

\begin{itemize}
    \item Using embeddings of only the user's utterances within the retriever for demonstration selection yields better performance than including both user and agent utterances. This suggests that user input is a more reliable indicator of the underlying dialogue state.
    \item Embedding models trained on multilingual data (LaBSE) or specifically for dialogue tasks (D2F) achieve comparable results in our setup, indicating that domain-specific fine-tuning of the embedding model may not be necessary for achieving good performance in our ICL-based DST system.
    \item We also find that tags indicating speaker roles (user, agent), the number of demonstrations, and dialogue history in the demonstrations play an important role in the final DST performance.
\end{itemize}

\begin{figure*}[!t]
    \centering
    \begin{tikzpicture}
	[
	rect/.style={minimum width=23mm,minimum height=1cm,text width=23mm,
		align=center, rectangle,draw,thick, rounded corners, fill=gray!20},
	rect2/.style={minimum width=23mm,minimum height=1cm,text width=23mm,
		align=center, rectangle,draw,thick, rounded corners, fill=white},
        var/.style={minimum width=20mm,minimum height=1.2cm,text width=20mm,
		align=center,rectangle,align=center},
        arr/.style={->,>=stealth',semithick},
        ]
        \node (t) [var] at (-2.8, 0) {User turn \\ (test sample)};
	\node (r) [rect]  at (0, 0)  {Demonstration \\ retriever \\ (embedding models)};
	\node (p) [rect]  at (3, 0)  {Prompt \\ constructor (dialogue history, speaker tags)};
        \node (l) [rect]  at (6, 0)  {Inference with LLM \\ (constrained decoding)};
        \node (pp) [rect2]  at (9, 0)  {Post-process \\ (JSON repair)};
        \node (o) [var] at (11.8, 0) {Predicted dialogue state};
        \draw [arr] (t) edge (r);
        \draw [arr] (r) edge (p);
        \draw [arr] (p) edge (l);
        \draw [arr] (l) edge (pp);
        \draw [arr] (pp) edge (o);
    \end{tikzpicture}
    \caption{Block diagram of the ICL scheme used in our experiments. Factors influencing the shaded parts are studied in this paper.}
    \label{fig:overall_system}
    \vspace{-0.3cm}
\end{figure*}
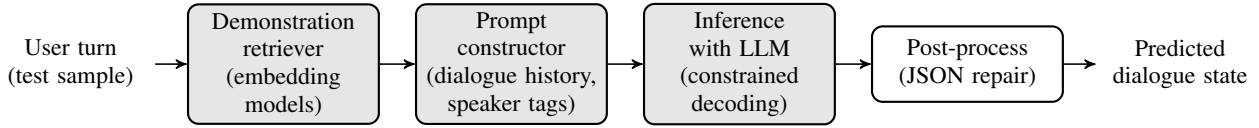

The remainder of this paper is organised as follows. Section \ref{sec:methodology} details our methodology and describes the key components involved. Section \ref{sec:experiments} describes our experiments and presents the results, showing the performance of our system under different configurations. Section \ref{sec:analysis} provides an in-depth analysis of our results and highlights the main findings of our investigation. Finally, Section \ref{sec:concl} concludes by summarising our contributions and outlining potential avenues for future work in the field of ICL-based DST.

\section{Methodology}
\label{sec:methodology}

This section provides a detailed step-by-step description of our proposed system. Figure \ref{fig:overall_system} provides a high-level representation of the data flow and the interactions between different modules of the system.

\subsection{Demonstration retriever}
The selection of appropriate demonstrations is crucial for effective in-context learning, as the quality and relevance of the demonstrations directly influence the LLM's ability to generalise to new examples.
While various strategies exist for selecting demonstrations, such as ensuring coverage across all domains or including multi-domain examples to enhance robustness, we observed that retrieving demonstrations from semantically similar dialogues yield better performance. Semantically similar dialogues are likely to contain relevant contextual information (domains) and slot keys that will be helpful for accurately tracking the dialogue state.

Our system employs a retriever that extracts embeddings of each dialogue turn of the user along with its history, using a pre-trained sentence embedding. In this work, we compare two embedding models. The first is LaBSE that was trained on multilingual parallel data, which efficiently captures cross-lingual features that are language agnostic \cite{feng-etal-2022-LaBSE}. The second is Dialog2Flow (D2F), which is trained with a contrastive objective to produce semantically similar embeddings for turns with similar dialogue acts \cite{burdisso-etal-2024-dialog2flow}.
To improve efficiency and reduce computational overhead, embeddings of the training set are pre-computed. During inference time, the retriever selects the $K$ \textit{closest} training samples to the test sample based on \textit{cosine similarity}.

\subsection{Prompt construction}
The prompt serves as the primary interface of the LLM and guides its output. Our prompt template consists of three modular blocks: the conversation history, the domain(s) identified in the conversation, and the corresponding slot key-value pairs for each domain. This modular design provides flexibility, allowing the template to accommodate long conversations, multiple domains, and respective slots while maintaining a structure that is easy to parse and interpret by the LLM. The demonstrations are ordered so that the most relevant example is closest to the test sample \cite{liu2022makes}. A simple instruction is given to the model at the beginning of the prompt. Fig.~\ref{fig:prompt_example} shows an example input prompt. The slots are represented using a JSON dictionary schema, as this allows for easy post-processing.

\subsection{Inference}
The prompt with the demonstration examples are passed as input the LLM, and we perform constrained decoding, i.e., decoding each slot value given the domain and slot key. Note that for a given domain (e.g. taxi), the possible slot keys (e.g. leaveBy, arrriveAt, destination, departure) are predefined in the schema.

\begin{figure}[htbp]
    \centering
    \raisebox{-0.5ex}{ 
    \begin{tcolorbox}[colback=gray!20, colframe=white]
        \begin{minipage}{\dimexpr\linewidth-2\fboxsep-2\fboxrule\relax}
            \textcolor{RawSienna}{Instruction: Identify the slot value.} \\
            \textcolor{olive}{\textbf{User:} Can you help me get a taxi to Pizza Hut Fen Ditton? \textbf{Agent:} Sure. Where do you want to depart from? \textbf{User:} I want to depart from Sidney, Sussex College, also I need a reservation there.}
            \textcolor{teal}{Domain: [``taxi", ``restaurant"]}
            \textcolor{brown}{Slots: \{``taxi": \{``arriveBy": ``not mentioned", ``departure": ``sidney sussex college", ``destination": ``pizza hut fenditton", ``leaveAt": ``not mentioned"\}, ``restaurant": \{``area": ``centre", ``day": "not mentioned", ``food": ``not mentioned", ``name": ``not mentioned", ``people": ``not mentioned", ``pricerange": ``expensive", ``time": ``not mentioned"\}\}}
            \hfill $\times K$ \\
            \textcolor{orange}{\textbf{User:} I would like a taxi from Saint John's College to Pizza Hut Fen Ditton. Domain: [``taxi"] Slots: \{``arriveBy": }
        \end{minipage}
    \end{tcolorbox}
    }
    \caption{\centering Illustration of the prompt format used for DST with simple a instruction. Demonstrations consist of a User-Agent turns (olive), domain(s) (teal), and corresponding slot values (brown). After $K$ such demonstrations, the test sample is presented in the same format (orange).}
    \label{fig:prompt_example}
\end{figure}


\subsection{Evaluation metric}
We evaluate the performance of our system using precision and recall of the predicted slot values. Precision measures the proportion of correctly predicted slots out of all predicted slots, while recall measures the proportion of correctly predicted slots out of all ground-truth slots.




\section{Experiments and results}
\label{sec:experiments}
In this section, we present the experimental setup and results obtained from our evaluation of the proposed system. We investigated the impact of various factors related to both the demonstration retriever and the prompt structure on overall DST performance. The primary goal of these experiments is to identify the factors that affect the tracking of \textit{the dialogue state} using in-context learning in LLM. All experiments consider the ground-truth domain when evaluating slot prediction performance.

\subsection{Dataset and pre-processing}
The MultiWOZ 2.4 dataset \cite{budzianowski-etal-2018-multiwoz,ye-etal-2022-multiwoz} is a multi-domain, multi-turn, task-oriented dialogue dataset comprising a total of 8438 dialogues, with 6438 in training and 1000 each in development and test sets. The dialogue conversations are pre-processed to make them suitable for the task of dialogue state tracking. Specifically, each turn is constructed by accumulating the current user turn with all preceding (user + agent) turns from the same dialogue, followed by the accumulated domain and slots. This yields 56,778 training samples (turns) and 7372 test samples (turns). The training set serves as the source for demonstration retrieval in our in-context learning framework.

\subsection{Large Language Models (LLMs)}

In this work, we focus mainly on the fully open-source OLMo-7B-Instruct model \cite{groeneveld-etal-2024-olmo}. Its transparency in terms of both model release and training dataset is a key consideration, aligning with principles of reproducible research. For comparison, we also include the open-weight models Mistral-7B-Instruct-v0.3 \cite{jiang2023mistral7b} and Llama3.2-3B-Instruct \cite{llama3herdmodels}. Unless explicitly stated, all the experiments are with OLMo-7B-Instruct.

\subsection{Factors related to the retrieved demonstrations}

\subsubsection{Choice of embedding extractor}
\begin{figure}
    \centering
    \includegraphics[width=0.98\linewidth]{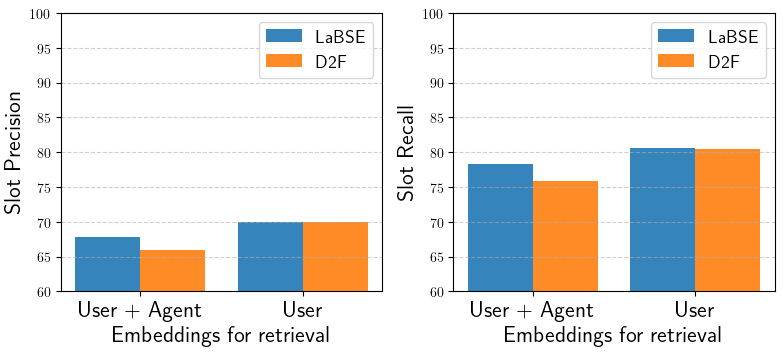}
    \caption{\centering Slot precision and recall for LaBSE and Dialog2Flow models when User-Agent and User only sentences are considered for embeddings for retrieval.}
    \label{fig:labse_d2f}
\end{figure}

Figure \ref{fig:labse_d2f} compares the DST performance of OLMo-7B-Instruct when using demonstration retrievers based on LaBSE and D2F. Here, embeddings are extracted for User-Agent turns and User-only turns. When User-Agent turns are used, LaBSE outperforms D2F. However, the performance of both models is comparable when only user turns are used for embeddings. This suggests that focussing on user utterances provides good semantic information for effective demonstration retrieval.

\subsubsection{Choice of dialogue turn for embeddings}
Here, we compare the performance of computing embedding using the user versus the user-agent turns. In both cases, the user-agent dialogue history is used as a context within the prompt. Table \ref{tab:U_emb_UA_dialog_history} presents the precision and recall scores for our system using a LaBSE based retriever. Computing embeddings using only the user's utterances consistently outperforms using the full user-agent dialogue turns, especially when using a larger number of demonstrations. This suggests that the user's utterances are more informative for capturing the relevant semantic information needed for effective demonstration retrieval.

\begin{table}[!t]
    \centering
    \caption{\centering Precision and recall for DST system relying on retriever based on LaBSE. Embeddings were extracted using either only User or User-Agent turns; while user-agent dialogue history was used in the context.}
    \label{tab:U_emb_UA_dialog_history}
    \begin{tabular}{crr|rr}
    \toprule
    \textbf{Max. demos} & \multicolumn{2}{c}{\textbf{Precision}} & \multicolumn{2}{c}{\textbf{Recall}} \\
     \textbf{}   &   \textbf{U} & \textbf{UA}    & \textbf{U}    & \textbf{UA}   \\ \midrule
    10  & 69.9  & 67.8  & 80.6  & 78.3 \\
    3   & 67.0 & 67.7   & 81.2  & 79.0 \\
    1   & 66.0 & 64.4   & 81.3  & 79.1 \\ \bottomrule
    \end{tabular}
\end{table}

\subsection{Factors related to the prompt structure}

\subsubsection{Speaker tags}
Speaker tags (``User:" and ``Agent:") are commonly used in dialogue systems to distinguish between user utterances and agent responses. This provides additional information to LLMs to classify domains and extract relevant slots. As we see in Table \ref{tab:spk_tag}, not having a speaker tag marginally improves precision but negatively affects recall.

\begin{table}[!t]
\centering
\caption{\centering Impact of speaker tags on DST performance of OLMo-7B-Instruct, Mistral-7B-Instructv0.3 and LLama3.2-3B-Instruct models.}
\label{tab:spk_tag}
\begin{tabular}{lcrr}
\toprule
\textbf{Model}  & \textbf{Speaker tag} & \textbf{Precision} & \textbf{Recall} \\ \midrule
OLMo-7BI        & Yes                  & 67.8          & \textbf{78.3}  \\
                & No                   &\textbf{ 68.1}  & 76.1   \\
                \midrule
Mistral-7BI     & Yes                  & 69.5        & \textbf{88.2}  \\
                  & No            & \textbf{70.2}    & 87.4  \\
\midrule
Llama3.2-3BI    & Yes                  & 67.5          & \textbf{87.5}  \\
                & No                   & 67.5               & 86.5  \\ \bottomrule
\end{tabular}
\end{table}

\subsubsection{Number of demonstrations}
The number of demonstrations included in the context can significantly impact the performance of ICL. Since the OLMo series models can accommodate a maximum of 2048 tokens, we make sure to select only the most relevant ones well within the token limit.  Table \ref{tab:demos} shows the performance comparison of DST systems employing different numbers of demonstrations, focussing on the User dialogue history configuration. We observe that performance generally improves with more demonstrations, up to a point. However, the optimal number of demonstrations may vary depending on the specific model and the complexity of the dialogue.


\begin{table}[!t]
    \centering
    \caption{\centering Performance comparison of DST systems employing different numbers of demonstrations across different LLMs. (LaBSE based retriever and User-only dialogue history).}
    \label{tab:demos}
    \begin{tabular}{clrr}
    \toprule
        \textbf{Max. demos} & \textbf{LLM} & \textbf{Precision} & \textbf{Recall}\\ \midrule
         & OLMo-7BI & 66.5 & 82.3\\
        1 & Mistral-7BI & \textbf{74.0} & 83.1\\
         & Llama3.2-3BI & 66.9 & \textbf{84.0}\\ \midrule
         & OLMo-7BI & 70.4 & 81.6\\
        3 & Mistral-7BI & \textbf{73.9 }& \textbf{83.1}\\
         & Llama3.2-3BI & 70.3 & 82.8\\ \midrule
         & OLMo-7BI & 71.6 & 80.0\\
        5 & Mistral-7BI & \textbf{74.3} & \textbf{83.0}\\
         & Llama3.2-3BI & 71.5 & 82.6\\ \midrule
         & OLMo-7BI & 73.2 & 79.3\\
        10 & Mistral-7BI & \textbf{74.8} & \textbf{83.1}\\
         & Llama3.2-3BI & 72.2 & 82.3\\ \bottomrule
    \end{tabular}
    \label{tab:my_label}
\end{table}



\subsubsection{Dialogue history in the prompt}
Here, we compare the impact of using user-agent turns or user-only utterances as the dialogue history within the prompt template. In both cases, the demonstrations are retrieved based on the respective type of sentences (user-agent or user-only). Table \ref{tab:dialog_history} presents the results of this comparison. It can be observed that, even with a small number of demonstrations (1 or 3), using User-only utterances in the dialogue history consistently yields better precision and recall than using User-Agent turns.

\begin{table}[]
\centering
\caption{\centering Impact of considering  user (U) vs. user-agent turns as dialogue history in the prompt template.}
\label{tab:dialog_history}
\begin{tabular}{crr|rr}
\toprule
  \textbf{Max. demos.} & \multicolumn{2}{c}{\textbf{Precision}} & \multicolumn{2}{c}{\textbf{Recall}} \\
\textbf{} & \textbf{U}   & \textbf{UA} & \textbf{U} & \textbf{UA} \\ \midrule
10  & 73.2      & 67.7      & 79.3    &   78.2    \\
3   &   70.4    &   67.0    &   81.6    & 79.0    \\
1   &   66.5    &   64.4    &   83.2    & 79.1    \\ \bottomrule
\end{tabular}
\vspace{-0.1cm}
\end{table}

\section{Analysis}
\label{sec:analysis}



\subsection{Relevance and coverage of demonstrations}
Here, we analyse the relevance and coverage of the selected demonstrations with respect to the slots that need to be predicted in the target instance.
To quantify the relevance and coverage of demonstrations, we compare the slot-keys (e.g. restaurant-name, restaurant-people, taxi-departure) present in the demonstrations to the ground truth slot-keys of the corresponding test sample. Specifically, we compute ``slot-key" precision and ``slot-key" recall, which can be interpreted as relevance and coverage, respectively.

\begin{figure}[!t]
    \centering
    \includegraphics[width=0.98\linewidth, height=6cm]{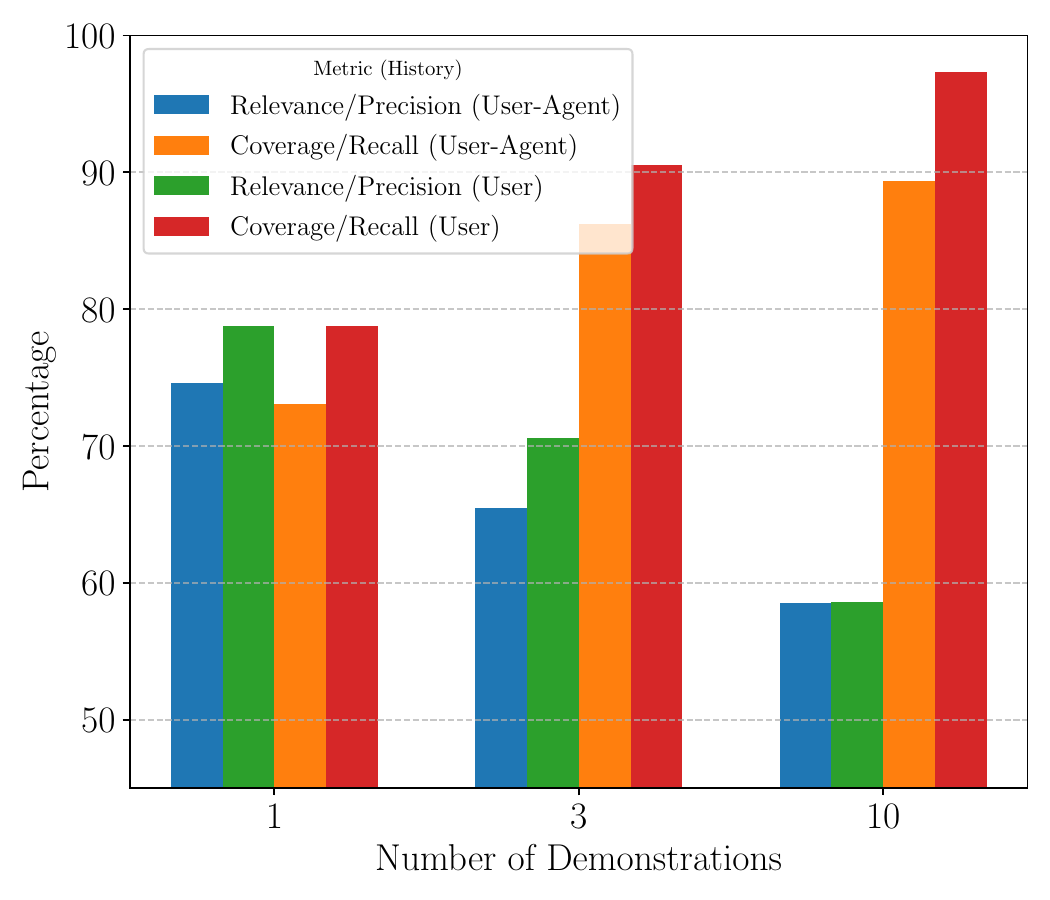}
    \caption{\centering Demonstration relevance and coverage for User-Agent and User only dialogue history in 1, 3, 10 number of demonstrations from LaBSE based retriever.}
    \label{fig:labse_relevance_comparison}
    \vspace{-5pt}
\end{figure}

\begin{figure}[!t]
    \centering
    \includegraphics[width=0.98\linewidth, height=6cm]{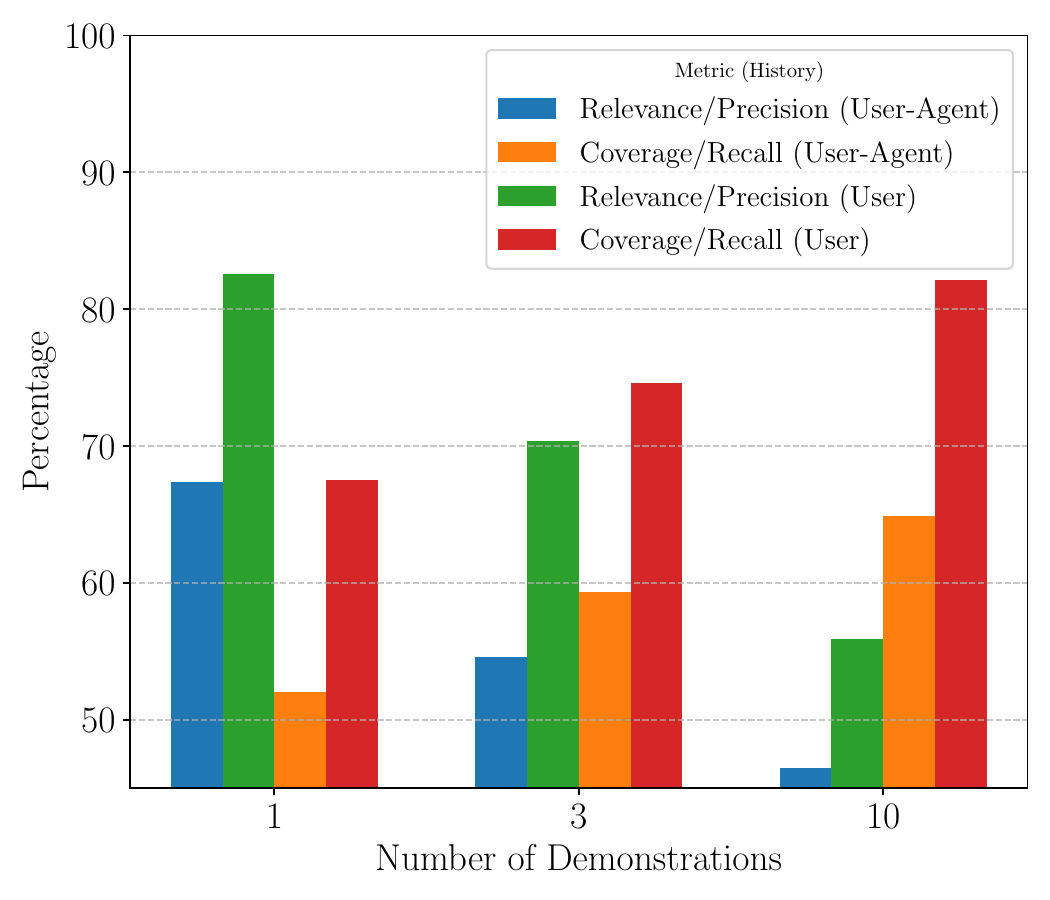}
    \vspace{-5pt}
    \caption{\centering Demonstration relevance and coverage for User-Agent and User only dialogue history in 1, 3, 10 number of demonstrations from D2F based retriever.}
    \label{fig:d2f_relevance_comparison}
    \vspace{-5pt}
\end{figure}

Figures \ref{fig:labse_relevance_comparison} and \ref{fig:d2f_relevance_comparison} illustrate the slot relevance and coverage for demonstrations retrieved using LaBSE and D2F embeddings, respectively. From these figures, we observe that slot relevance tends to decrease as slot coverage increases. A notable observation is that demonstrations retrieved using LaBSE exhibit better slot relevance and coverage compared to those retrieved using D2F.

The effects of the relevance and coverage of the demonstration can be observed in Table \ref{tab:emb_model}, which presents the precision and recall predicted for different configurations.

\begin{table}[]
\centering
\caption{\centering Performance comparison of max. number of demonstrations with LaBSE and D2F based retrievers.}
\label{tab:emb_model}
\scalebox{0.85}{
\begin{tabular}{crr|rr}
\toprule
 \textbf{Max. number of} & \multicolumn{2}{c}{\textbf{Precision}} & \multicolumn{2}{c}{\textbf{Recall}} \\
 \textbf{demonstrations} & LaBSE & D2F & LaBSE & D2F \\
 \midrule
    10  & 73.2 & 73.2  & \textbf{79.3} & 78.9\\
    3   & \textbf{70.3} & 68.4  & 81.5 & \textbf{81.6}6 \\
    1   & \textbf{66.5} & 66.1  & 83.2 & 83.2 \\ \bottomrule
\end{tabular}
}
\end{table}

\subsection{Decoding strategy}
We evaluate the effectiveness of the model using two distinct approaches: (1) prediction of the slot value when the slot key is provided as input and (2) generation of the entire pair of slot key value. Although slot key-value generation achieves higher precision, it exhibits a significantly lower recall due to a higher error rate with respect to the ground truth.

\begin{table}[!t]
    \centering
    \caption{\centering Performance Comparison of Decoding Strategies for Dialogue State Tracking using OLMo-7BI model.}
    \label{tab:decoding strategy}
    \begin{tabular}{lrr}\toprule
        \textbf{Decoding strategy} & \textbf{Precision} & \textbf{Recall}\\ \midrule
        Slot value prediction (given key) & 67.8 & \textbf{78.3}\\
        Slot key-value generation & \textbf{69.2} & 53.3\\ \bottomrule
    \end{tabular}
    \vspace{-10pt}
\end{table}

\section{Conclusion}
\label{sec:concl}
In this paper, we systematically studied various factors that influence the \textit{in-context learning} abilities of LLMs for dialogue state tracking. We found that (a) demonstration retrievers based on a general-purpose embedding model such as LaBSE perform as good as dialogue-specific modes like D2F when using a maximum of 10 demonstrations. (b) User turns are better candidates for retrieval than user-agent turns. (c) Speaker tags have a minor but significant effect on the precision and recall across the three LLMs studied. (d) Three or more demonstrations do not yield significantly better results. (e) LaBSE and D2F based retrievers yield distinctive examples, where LaBSE has more slot relevance and coverage and hence results in better DST performance when having fewer demonstrations. We believe that our study helps us to understand the ICL abilities of LLMs for DST.


\pagebreak
\section{Acknowledgements}
The work was supported by European Union’s Horizon Europe project No. SEP-210943216 ``ELOQUENCE'', Czech Ministry of Interior project No. VK01020132 ``112'', European Defence Fund project ARCHER, and by Czech Ministry of Education, Youth and Sports (MoE) through the OP JAK project ``Linguistics, Artificial Intelligence and Language and Speech Technologies: from Research to Applications'' (ID:CZ.02.01.01/00/23\_020/0008518). Computing on IT4I supercomputer was supported by MoE through the e-INFRA CZ (ID:90254).
\bibliographystyle{IEEEtran}
\bibliography{mybib}

\end{document}